\pdfoutput=1

\documentclass[11pt]{article}

\usepackage[]{emnlp2021}
\usepackage{xcolor, soul}
\sethlcolor{green}

\usepackage{times}
\usepackage{latexsym}
\usepackage{enumitem}
\usepackage{xcolor}
\usepackage[export]{adjustbox}

\usepackage[T1]{fontenc}

\usepackage[utf8]{inputenc}

\usepackage{microtype}

\title{Unpacking the Interdependent Systems of Discrimination: \\Ableist Bias in NLP Systems through an Intersectional Lens}

\author{Saad Hassan{\normalfont{\textsuperscript{1}}} \and Matt Huenerfauth{{\normalfont\textsuperscript{1,2}}} \and Cecilia Ovesdotter Alm{{\normalfont{\textsuperscript{1,3}}}} \\

{\textsuperscript{1}Golisano College of Computing and Information Sciences} \\
{\textsuperscript{2}School of Information}
{\textsuperscript{3}College of Liberal Arts} \\
        Rochester Institute of Technology \\ Rochester, New York, 14623 USA \\
        \texttt{\{sh2513,matt.huenerfauth,coagla\}@rit.edu}}

\date{}

\begin{document}
\maketitle
\begin{abstract}

Much of the world's population experiences some form of disability during their lifetime. Caution must be exercised while designing natural language processing (NLP) systems to prevent systems from inadvertently perpetuating ableist bias against people with disabilities, i.e., prejudice that favors those with typical abilities. We report on various analyses based on word predictions of a large-scale BERT language model. Statistically significant results demonstrate that people with disabilities can be disadvantaged. Findings also explore overlapping forms of discrimination related to interconnected gender and race identities.
\end{abstract}

\section{Introduction}
Over one billion people experience some form of disability \cite{WHO}, and 25\% of U.S. adults live with some disability \cite{cdc_2018}. Several studies have shown that people with disabilities experience discrimination and lower socio-economic status \cite{vanpuymbrouck2020explicit,nosek2007pervasiveness, szumski2020attitudes}. Recent studies have shown that biases against people with disabilities manifest in complex ways which differ from biases against other groups \cite{liasidou2013intersectional}. Although the intersection of disability, race, and gender has been understudied, recent research has stressed that the identities of people with disabilities should be understood in conjunction with other identities, e.g., gender \cite{caldwell2010we} or race \cite{frederick2019race, artiles2013untangling}, rather than considered fixed and gauged by atypical physical or psychological abilities. Despite increasing research on AI fairness and how NLP systems project bias against various groups \cite{blodgett-etal-2020-language,mccoy1998interface,emil2020towards,lewis2020implications,chathumalispeecur,borkan2019nuanced,bender2018data}, less attention has been given to examining systems' bias against people with disabilities \cite{trewin2018ai}.

Designing accessible and inclusive NLP systems requires understanding nuanced conceptualizations of social attitudes and prejudicial stereotypes that may be represented in learned models and thereby impact applications. For instance, hate-speech detection for moderating social-media comments may erroneously flag comments that mention disability as toxic \cite{hutchinson-etal-2020-social}. To better understand disability bias in NLP systems such as BERT, we build on prior work \cite{hutchinson-etal-2020-social} and additionally assess model bias with an intersectional lens \cite{jiang-fellbaum-2020-interdependencies}. The contributions are (1) examining ableist bias and intersections with gender and race bias in a commonly used BERT model, and (2) discussing results from topic modeling and verb analyses.

Our research questions are:

\begin{enumerate}[leftmargin=2.5\parindent]
    \item[\textbf{RQ1}] Does a pre-trained BERT model perpetuate measurable ableist bias, validated by statistical analyses?
    \item[\textbf{RQ2}] Does the model's ableist bias change in the presence of gender or race identities?
\end{enumerate}

\section{Background and Related Work}
There is a growing body of sociology literature that examines bias against people with disabilities and its relationship with cultural and socio-political aspects of societies \cite{barnes2018theories}. Sociological research has also moved from drawing analogies between ableism and racism to examining their intersectionality \cite{frederick2019race}. Disability rights movements have stimulated research exploring the gendered marginalization and empowerment of people with disabilities. The field of computing is still lagging behind. Work on identifying and measuring ethical issues in NLP systems has only recently turned to ableist bias---largely without an intersectional lens. While ableist bias differs, prior findings on other bias motivate investigation of these issues for people with disabilities \cite{spiccia2015word,blodgett-etal-2020-language}. 

There is a need for more work that deeply examines how bias against people with disabilities manifest in NLP systems through approaches such as critical disability theory \cite{hall2019critical}. However, a growing body of research on ethical challenges in NLP reveals how bias against protected groups permeate NLP systems. To better understand how to study bias in NLP, we focus on prior work in three categories: (1) observing bias using psychological tests, (2) analyzing biased subspaces in text representations such as word embeddings, and (3) comparing performance differences of NLP systems across various protected groups.

\begin{table*}[t]
\small
\centering
\begin{tabular}{|l|l|}
\hline
Blank            & \multicolumn{1}{c|}{Words or phrases used} \\ \hline
Disability & \begin{tabular}[c]{@{}l@{}}deaf person, blind person, person in a wheelchair, person with cerebral palsy, person with epilepsy,  \\ person who is chronically ill,  person with a mental illness, person with spinal curvature, \\  short-statured person, person with dyslexia, person with Downs syndrome, without a disability \\ \end{tabular} \\ \hline
Gender Identity & \begin{tabular}[c]{@{}l@{}}lesbian, gay, bisexual, queer, intersex, asexual, agender, androgyne, bigender, gender expansive,\\ genderfluid, genderqueer, nonbinary, polygender, transgender, trans, two spirit\end{tabular}  \\ \hline
Race$_{[\begin{small}\ref{fnlabel}\end{small}]}$ & American Indian, Asian, Black, Hispanic, Pacific Islander, White \\ \hline
Connecting Verbs & \begin{tabular}[c]{@{}l@{}}does, has, innovates, produces, develops, teaches, instructs, manages, leads, supervises, guides, \\ advises, feels, perceives\end{tabular}                       \\ \hline
\end{tabular}
\caption{Lexicon in template slots for creating sentence fragments to feed BERT and predict a subsequent word. The template ensured end of sentence after the predicted word. \emph{A person} was also used with connecting verbs.}
\descriptionlabel{}
\label{tab:slotslexicon}
\end{table*}

Research has sought to quantify bias in NLP systems using \textbf{psychological tests}, such as the  Implicit Association Test (IAT) \cite{greenwald1998measuring}, which can reveal influential subconscious associations or implicit beliefs about people of a protected group and their stereotypical roles in societies. Some work has studied correlations between data on gender and professions and the strengths of these conceptual linkages in word embeddings \cite{caliskan2017semantics, garg2018word}. Findings suggest that word embeddings encode normative assumptions, or resistance to social change, which can have implications for computational systems.

Analyzing \textbf{subspaces in text representations} like word embeddings can reveal insights about NLP systems that use them \cite{may2019measuring, chaloner2019measuring}. For example, \citet{bolukbasi2016man} developed a support vector machine to identify gender subspace in word embeddings and then identified gender directions by making ``gender-pairs (\textit{man-woman, his-her, she-he})''. They identified eigenvectors that capture prominent variance in the data. This work has been extended to include non-binary gender distinctions \cite{manzini2019black}. Researchers have also explored contextualized word embeddings bias at the intersection of race and gender. \citet{guo2020detecting} proposed methods for automatically identifying intersectional bias in static word embeddings. But debiasing has limitations. For example, \citet{gonen-goldberg-2019-lipstick} pointed out that even after attempting to reduce the projection of words on
a gender direction, biased/stereotypical words in the neighbors of a given word embedding remain \cite{gonen-goldberg-2019-lipstick}.

Other work has measured \textbf{performance bias of NLP systems} when used by someone from a protected group or when the input data mentions a protected group. Unfortunately, state-of-the-art systems pass on bias to other tasks. For example, a recent study found that BERT can perpetuate gender bias in contextualized word embeddings \cite{costa2020proceedings}. Some work has explored the effect on performance measures in NLP systems after replacing (swapping) majority-minority lexicons \cite{zhao2018learning, lu2020gender, kiritchenko2018examining}. Additionally, standard evaluation metrics  usually fail to take bias into account, nor are datasets carefully designed to reveal bias effects. Researchers have explored the utility of performance metrics for capturing differences due to bias and proposed new metrics \cite{dixon2018measuring, park2018reducing}. A recent systematic review raised this concern and pointed to datasets that probe gender bias \cite{sun2019mitigating}. There is a pressing need to develop \textbf{metrics, evaluation processes}, and \textbf{datasets} able to quantitatively assess ableist biases in NLP systems. As a first step, we critically assess how ableist biases manifest in NLP models and examine intersections of bias.

\section{Methods}

\begin{table*}[]
\centering
\small
\begin{tabular}{|c|c|c|c|c|c|c|}
\hline
Set & Disability & Gender  & Race    & Number of Sentences & Avg. Sentiment Score & Variance \\ \hline
A     &      &   &  & 14  & -0.013                & 0.004  \\ \hline
B     & Present    &   &  & 168  & -0.088                & 0.040  \\ \hline
C     & Present    &   & Present & 1008  & -0.080                & 0.041  \\ \hline
D     & Present    & Present &   & 2856  & -0.088                & 0.045  \\ \hline
E     & Present    & Present & Present & 17136   & -0.030                & 0.017  \\ \hline
\end{tabular}
\caption{One-way ANOVA followed by t-tests with Bonferroni corrections revealed a significant difference in the average sentiment for sets of referents. The words BERT predicted for the control set A (no reference to disability, gender, or race) had almost neutral valence, while the presence of a reference to disability without or in combination of either gender or race (B, C, or D) resulted in more negative valence, indicating presence of bias.}
\vspace{-0.5cm}
\label{tab:sentencesAE}
\end{table*}

\begin{table*}[h]
\centering
\small
\begin{tabular}{|c|l|}
\hline
Set & \multicolumn{1}{c|}{Topic names and top-k words}  \\ \hline
C     & \begin{tabular}[c]{@{}l@{}}
\underline{Unique words}: hair, objects, death, teach, safe, technologies, died, two, books, another \\
\underline{Topic C1}: something, pain, well, better, good, technology, fear, guilty, right, eyes, safe, film, books, objects\\
\underline{Topic C2: one, ass, children, died, two, death, sex, dead, light, ability, shit, called, fat, deaf}
\end{tabular} \\ \hline
D  & \begin{tabular}[c]{@{}l@{}}
\underline{Unique types}: play, failed, got, gas,  lost, words, nervous, teacher, movement, love\\
\underline{Topic D1}: light, others, objects, technology, eyes, hating, movement, one, self, skell, color, white, rod, gay\\\underline{Topic D2}: sex, safe, water, never, fire, oath, alive, two, nothing, good, guilty, work, drugs, anything\\ \end{tabular}   \\ \hline
E & \begin{tabular}[c]{@{}l@{}}
\underline{Unique words}: men, right, muscles, self, breast, oral, gender, bible, light, lead\\
\underline{Topic E1}: something, blood, safe, fire, white, alive, eating, guilty, color, fear, considered, heard, hip, pain\\ \underline{Topic E2}: children, reading, pain, movement, able, water, using, died, teach, black, called, disability, two, good\end{tabular} \\ \hline
\end{tabular}\label{tab:topicmodeling}
\caption{From sets C, D, and E (which contained race and/or gender, in addition to disability), 10 unique words are shown that appeared in multiple Hierarchical Dirichlet topics for that set (but not in the topics of any other set).  Word lists from two example topics for each set are also shown.  Some topics and predicted set-specific words are notably negative (\emph{death, drugs, failed, fear, guilty, hating, lost, pain}).}\label{tab:topicmodeling}
\end{table*}

We build on the work of \citet{hutchinson-etal-2020-social} which used a fill-in-the-blank analysis--originally proposed by \citet{kurita-etal-2019-measuring}--to study ableist bias in pre-trained BERT representations. 
We used BERT large model (uncased),  a pretrained English language model \cite{devlin-etal-2019-bert}. We adjusted their analysis method to examine ableist bias together with gender and racial bias. Our analysis method involves creating \textbf{template sentence fragments} of the form \emph{The [blank1] [blank2] [blank3] person [connecting verb] <predicted using BERT>}. The slots (blank1, blank2, blank3) were filled in based on three lists with referents related to disability, gender, and race. The disability list was provided by \citet{hutchinson-etal-2020-social}.\footnote{The list was adapted from Table 6 in \citet{hutchinson-etal-2020-social} containing recommended phrases used for referring to disability, which had been used in their bias analysis. Their list also included a phrase \textit{"a person without a disability."}} The list for race included the five categories in the U.S. census \cite{BUR2020}\footnote{For the set of race or ethnicity referents, using census terminology, one term was selected if two were provided (except for \textit{two or more races}) and \textit{Hispanic} was also included. \label{fnlabel}}, and the list for gender was based on guidelines for gender inclusiveness in writing  \cite{bamberger2021language}. The three slots before the connecting verb were systematically completed with combinations of ${0-3}$ race ([blank1]), gender identity ([blank2]) and disability ([blank3]) referents. BERT predicted text after the verb. The final set included \textbf{21,182 combinations of disability, gender, race, and connecting verbs}. The referents used are in Table  \ref{tab:slotslexicon}.

 Analysis was restricted to the 5 sets of sentences in Table \ref{tab:sentencesAE}, which also shows the number of sentences per set. Sets B-E included disability referents with or without gender or race referents. The connecting words included frequent verbs (e.g., \emph{does, has}), but also verbs with more semantic content (e.g., \emph{develops, leads}) to ensure a holistic and less verb-dependent analysis. A subsequent one-way ANOVA test motivated averaging results for connecting words in subsequent analysis. For each verb, we also used a baseline sentence of the form \emph{The person \textit{[connecting verb]} \textit{\textless predicted using BERT\textgreater}}, as a control set A. To  quantitatively and qualitatively uncover bias in the  sets, we performed \textbf{sentiment analysis} and \textbf{topic modeling}.

Following \citet{hutchinson-etal-2020-social} and \citet{kurita-etal-2019-measuring}, BERT was trained to predict the masked word. 
Each sentence fragment was input ten times, resulting in 10 predicted words (without replacement) per stimulus. Given the added number of referents and connecting words, a three step filtering process was performed where BERT output was carefully inspected, and nonsensical, ungrammatical output was manually filtered out in context.

\begin{enumerate}
    \item We removed any predicted punctuation tokens resulting in incomplete sentences.
    \item We removed predicted function words resulting in ungrammatical sentences.
    \item If still needed, in very few cases, removal of repeated or blank output, e.g., \textit{The Hispanic intersex person in a wheelchair perceives perceives.}
\end{enumerate}

This sometimes resulted in fewer than 10 words for stimuli. In our final set of results, 83,268 out of 211,820 (21,182 sentences times 10 predicted words) remained. The dataset of sentences has been made available for research.\footnote{List of sentences is available at: \url{https://github.com/saadhassan96/ableist-bias}.\label{note}}

Each  predicted word not filtered out was added in a carrier sentence template \emph{The person [connecting verb] <BERT predicted word>} to obtain a sentiment score. The average sentiment score for each of the five combinations of sets of referents to disability, gender, race or no referent (Table \ref{tab:sentencesAE}) were computed using the sentiment analyzer of the Google cloud natural language API \cite{GoogleNLP}. The sentiment scores ranged between -1.0 (negative) and 1.0 (positive) and refers to the overall emotional valence of the sentence. For example, the -0.088 average sentiment score of set B in Table \ref{tab:sentencesAE} would be weak-negative to neutral.

After confirming statistical normalcy with the Shapiro-Wilk test \cite{razali2011power}, one-way analysis of variance (ANOVA) examined differences in set averages \cite{cuevas2004anova} since there were multiple sets and their sentence counts differed. Post-hoc pairwise comparisons examined significant differences of sets \cite{armstrong2014use}. 

Additionally, after the same filtering,
the Hierarchical Dirichlet process, an extension of Latent Dirichlet Allocation \cite{jelodar2019latent}, was used on the BERT predicted output per set to discover abstract topics and  words associated with them. This non-parametric Bayesian approach clusters data and discovers the number of topics itself, rather than requiring this as an input parameter \cite{asgari2017utility,teh2010hierarchical}.

\begin{figure*}
\vspace{-0.35cm}
\centering
  \includegraphics[width=0.8\textwidth,height=7cm,cfbox=black 1pt]{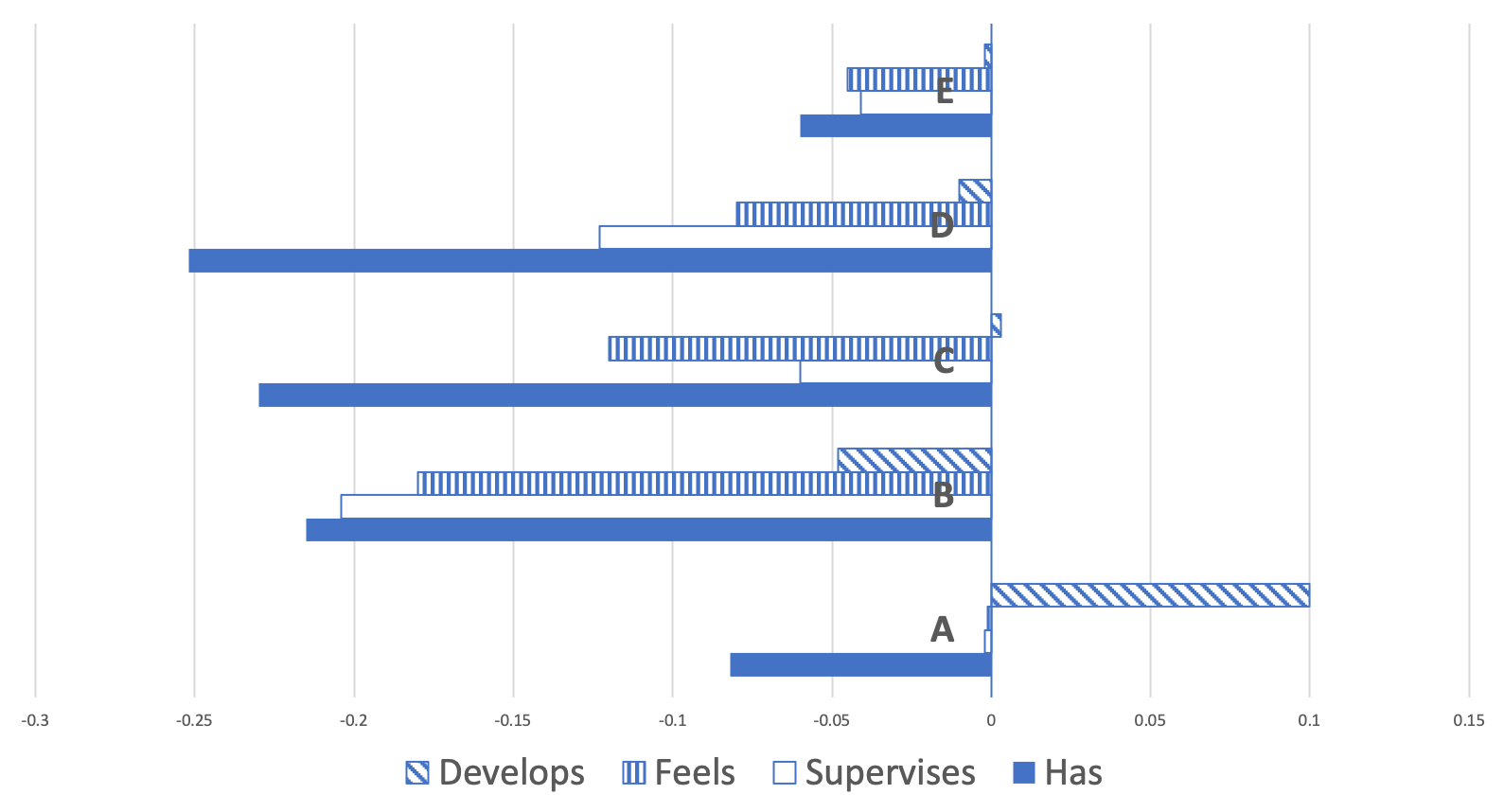}
  \caption{
  Averaged sentiment for selected connecting verbs \emph{develops, feels, supervises} and \textit{has}. For control set A, verbs have near-neutral sentiment, aside from \textit{has} (negative) and \textit{develops} (positive). In contrast, set B (disability) and sets C and D (disability and gender or race) are negative. Per-verb differences include, e.g., \emph{supervises} is most negative for set B, \emph{has} most negative for set D, and \emph{feels} slightly more negative for set C than set D.
  }
  \label{fig:perverbanalysis}
  \vspace{-0.5cm}
\end{figure*}

\section{Results and Discussion}
The average sentiment score in sentences that mentioned disability (with or without other sources of biases) was -0.0409 (weighted average of sets B, C, D, and E) which is more negative than sentiment score for sentences that did not mention disability -0.0133. Table \ref{tab:sentencesAE} shows the number of sentences in each set of sentences A-E, and the sets' average sentiment scores and variance. One-way ANOVA showed that the effect of choice of referents in sentences used for BERT word prediction was significant (F= 116.0 , F crit. = 2.372, p = $5.21^{-98}$). Post hoc analyses using t-test with Bonferroni corrections showed 6 out of 10 pairs as significantly different: A vs. B, A vs. C, A vs. D, B vs. E, C vs. E, D vs. E. Other pairs were not: A vs. E, B vs. C, B vs. D, and C vs. D. The findings reveal that sentence sets mentioning disability (alone or in combination with gender or race) are more negative on average than control sentences in set A. Set E's average sentiment appears less negative which may relate to this set's much higher sentence count. Figure \ref{fig:perverbanalysis} exemplifies set A's near-neutral sentiment and also that there are per-verb sentiment differences. Select topic output for intersectional sets in Table \ref{tab:topicmodeling} indicates negative associations for several predicted words.   

NLP models are deployed in many contexts and used by people with diverse identities. Word prediction is used for automatic sentence completion \cite{spiccia2015word}, and it is critical that it does not perpetuate bias. That is, it is insensitive to predict words with negative connotation given referents related to disability, gender, and race.  Our findings reveal ableist bias in a commonly used BERT language model. This also held for intersections with gender or race identity, reflecting observations in sociological research \cite{ghanea2013intersectionality,kim2020intersectional}. The average sentiment for set A was significantly lower than for the combination of other sets, affirming \textbf{RQ1}. Pairwise comparisons of set A with sets B, C, and D showed significant differences. The average sentiment of set A was also smaller than set E but not significantly.

The answer to \textbf{RQ2} is more nuanced. Results suggest similar sentiment for combining disability with race and gender, though per-verb sentiment analysis indicates it would be beneficial to explore a larger vocabulary for sentence fragments, and combine quantitative measures with deeper qualitative analysis. We begin to explore the utility of topic modeling by examining topics or unique words in vocabulary generated by BERT for sentence fragment sets.

Our findings have implications for several NLP tasks. Hate-speech or offensive-content detection systems on social media could be triggered by someone commenting neutrally about topics related to disability \cite{schmidt-wiegand-2017-survey}. Automatic content filtering software for websites may wrongly determine that keywords related to disability topics should be a basis for filtering, thereby restricting access to information about disability topics \cite{10.1145/3232676}. 
Further, ableist biases can have an impact on the accuracy of automatic speech recognition when people discuss disabilities if language models are used. It could also impact text simplification that is NLP-driven. These results could also be important if NLP models are used for computational social science applications.

Our findings also speak to the prior research on analyzing intersectional biases in NLP systems. Intersectionality theory posits that various categories of identities overlay on top of each other to create distinct modalities of discrimination that no single category shares. Prior work had examined this in the context of race and gender, e.g., \citet{lepori-2020-unequal} examined bias against black women who are represented in word embeddings as less feminine than white women. To the best of our knowledge our paper was also the first to conduct an analysis of intersectional ableist bias using different verbs. The complements likely to follow actions verbs like those in our study, e.g. innovates, leads, or supervises, may depend upon inadvertently learned stereotypes about the subject of each verb. Our analysis of these predictions helps to reveal such bias and how it may manifest in social contexts.  

\section{Conclusion and Future Work}
Our findings reveal ableist biases in an influential NLP model, indicating it has learned undesirable associations between mentions of disability and negative valence. This supports the need to develop metrics, tests, and datasets to help uncover ableist bias in NLP models. The intersectionality of disability, gender, and race deserves further study.

This work's limitations are avenues for future research. We only studied the intersections of  disability, gender, and race. We did not explore race and gender, or their combination, without disability. Studies can also look at other sources of bias such as ageism and expand the connecting verbs. Our sentiment analysis was also limited to template carrier sentences with one word predicted by BERT. Future work can allow BERT or other language models to predict multiple words and analyze the findings. We focused on a small number of manually selected verbs while comparing averaged sentiment. Future work could investigate a greater variety of verbs, and it could analyze more specifically how particular combinations of identity characteristics and verbs may reveal forms of social bias. 
For our analysis, we primarily used an averaged sentiment score. Future research can consider using other approaches to examine bias as well. Finally, future work can modify or improve different state-of-the-art debiasing approaches to remove intersectional ableist bias in NLP systems.

\bibliography{anthology,custom}
\bibliographystyle{acl_natbib}




\end{document}